\documentclass[11pt]{article}

\usepackage[preprint]{acl}

\usepackage{times}
\usepackage{latexsym}

\usepackage[T1]{fontenc}
\usepackage{iftex}
\ifPDFTeX
\usepackage[utf8]{inputenc}
\fi

\usepackage{microtype}
\usepackage{inconsolata}
\usepackage{graphicx}
\usepackage{rotating}
\usepackage{algorithm}
\usepackage{algpseudocode}
\usepackage{booktabs}
\usepackage{multirow}
\usepackage{enumitem}
\usepackage{caption}
\usepackage{subcaption}
\usepackage{amsmath}
\usepackage{amssymb}
\usepackage{etoolbox}

\AtBeginEnvironment{tabular}{\small}

\newcommand{\EAL}{EAL}
\newcommand{\real}{\textsc{REAL}}
\newcommand{\slk}{static LK}
\newcommand{\ealp}{\real{} \text{w/o RB}}
\newcommand{\ealb}{\real}
\newcommand{\wtv}{WTV}
\newcommand{\ealv}{\textsc{REAL-V}}
\newcommand{\ealk}{\textsc{REAL-K}}
\newcommand{\dpace}{D-PACE}
\newcommand{\dblast}{DBLast}
\newcommand{\vsd}{VSD}
\newcommand{\dflash}{DFlash}
\newcommand{\domino}{Domino}
\newcommand{\dspark}{DSpark}

\title{\real{} Loss: Optimizing \underline{R}eal \underline{E}xpected \underline{A}cceptance \underline{L}ength for Speculative Decoding}

\author{
  Yu Tian \\
  The University of Hong Kong \\
  \texttt{yutian23m@connect.hku.hk}
}

\begin{document}
\maketitle

\begin{abstract}
Speculative decoding accelerates LLM inference by having a draft model propose multiple tokens for the target model to verify in one pass.
The speedup is governed by the expected acceptance length (\EAL{}), yet standard training objectives such as cross-entropy, KL divergence, and total variation focus only on distribution matching or single-step overlap.
Recent studies instead propose block-level training objectives, aiming to optimize \EAL{} directly.
However, these objectives remain surrogates rather than exact formulations of \EAL{}.
For example, some average later acceptance over target-distributed continuations rather than the verification branches selected by accepted tokens, introducing a covariance gap that can reverse the gradient direction.
We analyze this gap and derive \real{}, a real \EAL{} training loss that uses path weights for unbiased estimation and Rao--Blackwellization for variance reduction.
In a controlled loss-level comparison with a Qwen3.6-35B-A3B target and a \domino{} drafter, \ealb{} outperforms \wtv{}, \vsd{}, DBLast, and the \slk{} baseline on probabilistic serving acceptance length across 9 task subsets and 5 evaluation seeds.
These gains require only a change to the training loss, with no modification to the drafter architecture or inference procedure, demonstrating the practicality of real \EAL{} training.
A reference implementation of the \real{} loss is available at \url{https://github.com/King-ty/speculators/tree/real-loss}.

\end{abstract}

\section{Introduction}
\label{sec:intro}

Speculative decoding accelerates large language model inference by letting a small draft model propose multiple tokens that the target model verifies in one forward pass \cite{leviathan2023fast,chen2023accelerating}.
At a fixed draft and verification configuration, its speedup is governed by the expected number of accepted draft tokens per round, known as the expected acceptance length (\EAL{}).
However, the dominant training objectives, such as cross-entropy and KL divergence, optimize token-level distribution matching rather than \EAL{} \cite{zhou2023distillspec}, causing a gap between the training objective and the inference metric.

\begin{figure}[t]
\centering
\includegraphics[width=0.98\columnwidth]{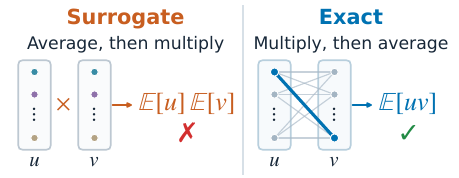}
\caption{Two-step illustration of branch dependence. The surrogate averages before multiplying, while the exact objective multiplies before averaging.}
\label{fig:branch-dependence}
\end{figure}

Acceptance-oriented objectives narrow this gap at different levels.
Single-step total variation losses optimize the one-step acceptance rate \cite{goel2024direct,samarin2026lk}, while recent work extends this idea to block-level objectives that aim to optimize \EAL{} directly \cite{zou2026variational,karimi2026dblast,li2026bebop,liu2026angelspec,xia2026acceptance}.
For example, the end-to-end TV loss \cite{li2026bebop} and the windowed total variation (\wtv{}) objective of \citet{xia2026acceptance} multiply per-step average acceptance rates over the proposal window.
However, this product form is a biased surrogate because an accepted token both extends the accepted prefix and determines the prefix used to verify the next draft token.
Multiplying marginal acceptance rates discards information about which tokens survive verification, so later acceptance is averaged over target-sampled paths rather than weighted by the propose-and-accept mass of the surviving branches, as shown in Figure~\ref{fig:branch-dependence}.
This mismatch appears as $\operatorname{Cov}(b_k,a_{k+1})$, where $b_k$ is the path weight under target sampling and $a_{k+1}$ is the next-step acceptance rate.
The covariance can have either sign and can reverse the gradient direction, causing the surrogate to point away from the true objective, as shown in Section~\ref{sec:method-gap}.
In practice, we find this covariance positive throughout training, consistent with the observed gap between the branch estimate and the surrogate (Section~\ref{sec:results-diagnostics}).

To address this gap, we derive an unbiased path estimator of the exact \EAL{} for chain speculative decoding with stochastic rejection sampling.
On trajectories sampled from the target, each sampled token is reweighted by a path weight $b_i = \min(1, q(y_i)/p(y_i))$ to recover the true propose-and-accept branch distribution.
We then obtain the branch estimator by marginalizing the final token in each term via Rao--Blackwellization, reducing per-term variance while preserving the expected gradient \cite{rao1945information,blackwell1947conditional}.
The branch estimator is differentiable with respect to the draft parameters, allowing direct backpropagation during training.
Finally, we derive our final training loss, \real{}, by adding an auxiliary CE term to stabilize training from scratch.

We extensively evaluate \real{} using a Qwen3.6-35B-A3B MoE target \cite{qwen36_35b_a3b}, a \domino{} drafter \cite{huang2026domino}, target-sampled data shared across methods, and a three-epoch training budget.
\ealb{} reaches 4.792 probabilistic acceptance length, outperforming all compared baselines (4.657--4.749) across 9 task subsets and 5 evaluation seeds, demonstrating the effectiveness of real \EAL{} estimation.
The variant without Rao--Blackwellization achieves comparable acceptance length at serving temperature $T=1$, while \ealb{} reduces aggregate gradient variance by 52--58\% and generalizes better across serving temperatures.
Pure CE and \ealb{} w/o CE also underperform \ealb{}, showing that the real \EAL{} term and the auxiliary CE term each contribute to the final result.
These gains require only a change to the training loss, with no modification to the drafter architecture or inference procedure, and improve average acceptance length across tasks and seeds, underscoring the practical significance of our method.

Our contributions are:
\begin{itemize}[leftmargin=1.2em,itemsep=0pt]
\item We identify a structural bias in existing block-level surrogates and derive a covariance diagnostic that links the discarded branch dependence to potential gradient reversal.

\item We derive \real{}, a real \EAL{} training loss based on an unbiased and differentiable branch estimator, using path weights to correct bias and Rao--Blackwellization to reduce variance, with changes limited to the training loss.

\item We show that \ealb{} improves probabilistic serving acceptance length over all baselines across 9 task subsets and 5 seeds, demonstrating the effectiveness of real \EAL{} estimation.
\end{itemize}

\section{Preliminaries}
\label{sec:background}

\subsection{Speculative Decoding Framework}
\label{sec:sd_framework}

We consider standard chain speculative decoding \cite{leviathan2023fast,chen2023accelerating,xia2023speculative} with stochastic rejection sampling, where \emph{chain} refers to the causal dependence between consecutive draft positions, allowing implementations with a parallel backbone and a causal head.
A frozen target model $p$ and a trainable draft model $q_\theta$ share a vocabulary $\mathcal{V}$.
Each decoding round starts from a fixed prefix $h$.
The draft proposes up to $\gamma$ tokens, which the target evaluates in parallel but accepts sequentially, stopping at the first rejection.
At step $i$, with prefix $h_i = (h, y_{<i})$, the draft proposes $y_i \sim q_{\theta,i}(\cdot \mid h_i)$, and the proposal is accepted with probability
\begin{equation}
\label{eq:alpha}
\alpha_i(y_i \mid h_i) = \min\!\left(1, \frac{p_i(y_i \mid h_i)}{q_{\theta,i}(y_i \mid h_i)}\right).
\end{equation}
Verification stops at the first rejection, after which the target emits a correction token and the round ends.
Let $L \in \{0, \dots, \gamma\}$ be the number of accepted draft tokens in a round, excluding the correction token.
The expected acceptance length is
\begin{equation}
\label{eq:tailsum}
J_\gamma(h,\theta) = \mathbb{E}[L] = \sum_{k=1}^{\gamma} P(L \ge k),
\end{equation}
where the second equality is the tail-sum identity.
Acceptance length is a proxy for speculative decoding speedup because longer accepted prefixes reduce target verification rounds per generated token at a fixed verification configuration.

We write $p_i(v)$ and $q_i(v)$ for $p_i(v \mid h_i)$ and $q_i(v \mid h_i)$, suppressing the prefix $h_i$ for brevity, where $v$ ranges over $\mathcal{V}$ and $y_i$ denotes the token sampled at position $i$.
The propose-and-accept mass of token $v$ at step $i$ is
\begin{equation}
\label{eq:mass}
m_i(v) = q_i(v)\, \alpha_i(v) = \min\{p_i(v), q_i(v)\},
\end{equation}
and summing over the vocabulary gives the single-step acceptance rate
\begin{equation}
\label{eq:tv}
\begin{aligned}
a_i
&= \sum_{v \in \mathcal{V}} m_i(v)
= \sum_{v \in \mathcal{V}} \min\{p_i(v), q_i(v)\} \\
&= 1 - \mathrm{TV}(p_i, q_i),
\end{aligned}
\end{equation}
where $\mathrm{TV}$ is the total variation distance.

A subtle but important point is the distribution of the accepted token.
Conditioned on acceptance, the token follows
\begin{equation}
\label{eq:r}
r_\theta(v \mid h_i, \text{accept}) = \frac{m_i(v \mid h_i)}{a_i(h_i)},
\end{equation}
which normalizes the propose-and-accept mass and equals neither $p_i$ nor $q_i$.
An objective that averages over target trajectories without correcting for this selection can therefore weight future branches incorrectly.

\subsection{The Windowed Total Variation Objective}
\label{sec:wtv_def}

The closest related work includes the end-to-end TV objective \cite{li2026bebop} and the windowed total variation (\wtv{}) objective of \citet{xia2026acceptance}, which share a product-of-marginals form.
\begin{equation}
\label{eq:wtv}
\widetilde{J}_{\mathrm{\wtv{}}} = \mathbb{E}_{y \sim p}\!\left[\sum_{k=1}^{\gamma} \prod_{i=1}^{k} a_i(h_i)\right].
\end{equation}
This objective multiplies per-step acceptance rates over the proposal window, approximating the survival probability without preserving the accepted-token distribution in Equation~\eqref{eq:r}.
Section~\ref{sec:method-gap} derives its exact two-step gap and shows that it can reverse the gradient direction.

\section{Method}
\label{sec:method}

\subsection{The Covariance Bias of \wtv{}}
\label{sec:method-gap}

We use a two-step example to illustrate what the \wtv{} surrogate discards.
An accepted token determines the prefix on which the next proposal is verified.
For a fixed prefix $h$ and a two-step window ($\gamma = 2$), Equation~\eqref{eq:tailsum} gives the exact objective
\begin{equation}
\label{eq:twostep}
\begin{aligned}
J_2 &= a_1 + \sum_{v} m_1(v)\, a_2(hv) \\
    &= a_1 + a_1 \mathbb{E}_{v \sim r_1}[a_2].
\end{aligned}
\end{equation}
Here $r_1(v)=m_1(v)/a_1$ is the accepted-branch distribution defined in Equation~\eqref{eq:r}.
The second term weights each continuation's acceptance rate by its propose-and-accept mass $m_1(v)$ in Equation~\eqref{eq:mass}.
For comparison, the \wtv{} objective in Equation~\eqref{eq:wtv} for $\gamma = 2$ reads
\begin{equation}
\label{eq:wtv2}
\begin{aligned}
\widetilde{J}_{\mathrm{\wtv{}},2}
&= \mathbb{E}_{v \sim p_1}\!\left[a_1 + a_1 a_2(hv)\right] \\
&= a_1 + a_1\, \mathbb{E}_{p_1}[a_2].
\end{aligned}
\end{equation}
Here $a_1$ is a constant because it is computed on the fixed prefix $h$ and does not depend on the sampled token $y_1$.
\wtv{} weights the second-step acceptance rate by $a_1 p_1(v)$ instead of $m_1(v)$.
The two objectives can therefore favor different continuations even when their first-step acceptance rates are equal.

We define the path weight of token $v$ at step $i$ as
\begin{equation}
\label{eq:b}
b_i(v) = \min\!\left(1, \frac{q_i(v \mid h_i)}{p_i(v \mid h_i)}\right),
\end{equation}
which satisfies the key identity $p_i(v \mid h_i)\, b_i(v) = m_i(v \mid h_i)$.
Taking the expectation of $b_1$ under $p_1$ gives
\begin{equation}
\label{eq:eb}
\mathbb{E}_{p_1}[b_1] = \sum_v p_1(v)\, b_1(v) = \sum_v m_1(v) = a_1.
\end{equation}
We can then rewrite Equation~\eqref{eq:twostep} as
\begin{equation}
\label{eq:twostep_b}
J_2 = a_1 + \mathbb{E}_{p_1}[b_1\, a_2].
\end{equation}
Comparing Equations~\eqref{eq:twostep_b} and~\eqref{eq:wtv2},
\begin{equation}
\label{eq:cov}
\begin{aligned}
J_2 - \widetilde{J}_{\mathrm{\wtv{}},2}
&= \mathbb{E}_{p_1}[b_1 a_2] - \mathbb{E}_{p_1}[b_1]\, \mathbb{E}_{p_1}[a_2] \\
&= \operatorname{Cov}_{p_1}(b_1, a_2).
\end{aligned}
\end{equation}
The surrogate replaces the paired average of the path weight $b_1$ and continuation acceptance rate by the product of their marginal averages.
The covariance can have either sign, so \wtv{} is neither an upper nor a lower bound in general.
This bias is not specific to $\gamma = 2$, as the same branch-weighting mismatch appears in every prefix product for longer windows, and adjacent covariances diagnose the resulting local branch dependence.

The difference can also reverse the gradient direction.
Consider a two-step example on the vocabulary $\{A, B\}$.
The first step has $p_1 = (0.4, 0.6)$ and $q_1 = (0.8, 0.2)$, so $a_1 = \min(0.4, 0.8) + \min(0.6, 0.2) = 0.6$.
The second step has $p_2(\cdot \mid hA) = (0.9, 0.1)$ and $p_2(\cdot \mid hB) = (0.1, 0.9)$, and the draft shares $q_2 = (\theta, 1-\theta)$ on both branches with $0.1 < \theta < 0.9$.
Then $a_2(hA) = 0.1 + \theta$ and $a_2(hB) = 1.1 - \theta$, and direct enumeration of the four paths gives
\begin{equation}
\label{eq:counterexample}
J_2 = 0.86 + 0.20\,\theta, \quad \widetilde{J}_{\mathrm{\wtv{}},2} = 1.02 - 0.12\,\theta.
\end{equation}

\begin{figure}[t]
\centering
\subcaptionbox{branch weights in step 2\label{fig:counterexample_a}}{\includegraphics[width=0.495\columnwidth]{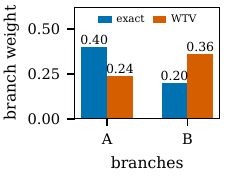}}\hfill
\subcaptionbox{opposite gradients\label{fig:counterexample_b}}{\includegraphics[width=0.495\columnwidth]{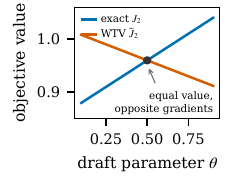}}
\caption{The two-step counterexample.
(a) The exact objective weights each branch by its propose-and-accept mass, while the \wtv{} surrogate redistributes that mass by target proportions.
(b) As a function of the shared draft parameter $\theta$, the two objectives have opposite gradients, coinciding in value at $\theta = 0.5$.}
\label{fig:counterexample}
\end{figure}

Accepted trajectories enter branch $A$ with mass $0.4$ and branch $B$ with mass $0.2$, so the true objective rewards increasing $\theta$, which improves the branch with more accepted mass in this example.
The \wtv{} surrogate redistributes the surviving mass by the target proportions $0.24$ and $0.36$ and rewards decreasing $\theta$ (Figure~\ref{fig:counterexample}).
At $\theta = 0.5$ the two objectives have equal value ($0.96$) but opposite gradients ($0.20$ vs $-0.12$).
The full enumeration and a second example where the surrogate reverses the ranking of two drafts are in Appendix~\ref{app:counterexample}.

Equation~\eqref{eq:cov} becomes a tool for detecting branch dependence: the covariance measures how much the path weight and continuation acceptance rate co-vary, and a nonzero value indicates that the \wtv{} surrogate misrepresents the branch structure.
Section~\ref{sec:results-diagnostics} examines this branch dependence and the gap between the exact and surrogate objectives during training.

\subsection{Unbiased Estimation via Path Weights}
\label{sec:method-ealp}

For a general window $\gamma$, the exact survival probability is
\begin{equation}
\label{eq:survival}
P(L \ge k \mid h) = \sum_{y_{1:k} \in \mathcal{V}^k} \prod_{i=1}^{k} m_i(y_i \mid h_i).
\end{equation}
Together with the tail-sum identity in Equation~\eqref{eq:tailsum}, this defines $J_\gamma$.
Each factor $m_i(y_i \mid h_i)$ is conditional on the preceding accepted tokens, so this expression preserves the branch structure without an independence assumption.

Direct evaluation of Equation~\eqref{eq:survival} is intractable because it sums over $|\mathcal{V}|^k$ paths.
Our estimators instead reuse continuations sampled from the frozen target, $y_{1:\gamma} \sim p$.
The derivation assumes that the trajectories are sampled from the target distribution $p$, while a draft-sampled variant would replace $p$ with $q$ and adjust the path weights accordingly.
On such a trajectory, computing the path weight $b_i = b_i(y_i)$ in Equation~\eqref{eq:b} requires only the sampled token's log-probabilities, which is much cheaper than directly evaluating Equation~\eqref{eq:survival}.

Let $W_0 = 1$ and $W_k = \prod_{i=1}^{k} b_i$.
Using $p_i(y_i\mid h_i)b_i=m_i(y_i\mid h_i)$ at every position gives $\mathbb{E}_{y \sim p}[W_k] = P(L \ge k)$.
This yields an unbiased estimator:
\begin{equation}
\label{eq:ealpath}
\widehat{J}_{\textsc{Path}} = \sum_{k=1}^{\gamma} W_k, \quad \mathbb{E}_{y \sim p}[\widehat{J}_{\textsc{Path}}] = J_\gamma.
\end{equation}
We call this the \emph{path estimator}.
However, its reliance on sampled tokens throughout the trajectory can introduce substantial variance.

\subsection{Variance Reduction via Conditional Marginalization}
\label{sec:method-ealb}
We observe that within each prefix term, the last factor has no later factors, so it can be marginalized exactly using the current vocabulary distributions.
Based on this observation, we further use Rao--Blackwellization \cite{rao1945information,blackwell1947conditional} to marginalize the last step of each term and reduce the variance of the path estimator.

Since $\mathbb{E}[b_k \mid h_k] = a_k(h_k)$, we can replace $b_k$ with its conditional expectation $a_k$:
\begin{equation}
\label{eq:ealbranch}
\widehat{J}_{\textsc{Branch}} = \sum_{k=1}^{\gamma} W_{k-1}\, a_k, \quad \mathbb{E}_{y \sim p}[\widehat{J}_{\textsc{Branch}}] = J_\gamma.
\end{equation}
We call this the \emph{branch estimator}.
For example, with $\gamma=3$, the path estimator reads $b_1 + b_1 b_2 + b_1 b_2 b_3$ while the branch estimator reads $a_1 + b_1 a_2 + b_1 b_2 a_3$, with only the last factor of each term replaced by its conditional mean.
Algorithm~\ref{alg:ealb} details the branch estimator computation.

\begin{algorithm}[t]
\caption{Computing the branch estimator.}
\label{alg:ealb}
\begin{algorithmic}[1]
\Require Frozen target $p$, draft $q_\theta$, window size $\gamma$
\State Sample trajectory $y_{1:\gamma} \sim p$
\For{$k = 1$ to $\gamma$}
  \State $b_k \gets \min(1, q_\theta(y_k \mid h_k) / p(y_k \mid h_k))$
  \State $W_k \gets \prod_{i=1}^{k} b_i$
  \State $a_k \gets \sum_v \min(p(v \mid h_k), q_\theta(v \mid h_k))$
\EndFor
\State $\widehat{J}_{\textsc{Branch}} \gets \sum_{k=1}^{\gamma} W_{k-1} a_k$
\State \Return $\widehat{J}_{\textsc{Branch}}$
\end{algorithmic}
\end{algorithm}

\paragraph{Variance guarantee.}
The conditional marginalization in the branch estimator is an instance of Rao--Blackwellization.
For each prefix contribution $k$, the marginalized gradient is the conditional expectation of its trajectory-based counterpart given $h_k$, and therefore has no greater covariance.
Summing the mean-squared errors of the prefix-gradient estimators gives the ordering $V_{\textsc{Branch}} \le V_{\textsc{Path}}$, where $V$ denotes the sum of gradient mean-squared errors across prefix contributions.
The reduction in this summed mean-squared error equals the gradient variance removed by marginalizing the final token of each prefix term.
Thus, the branch estimator preserves the exact expected gradient while eliminating this source of sampling noise from each prefix contribution.
A detailed proof is given in Appendix~\ref{app:variance-proof}, and the sampling variance reduction is measured at the main scale in Section~\ref{sec:results-ablation}.

\subsection{Gradient of \EAL{} Estimators}
\label{sec:method-gradient}
Both the path estimator and the branch estimator are differentiable with respect to the draft parameters $\theta$ wherever $a_k$ is differentiable.
Because the distribution $p$ does not depend on $\theta$, both estimators give $\nabla_\theta J_\gamma = \mathbb{E}_{y \sim p}[\nabla_\theta \widehat{J}]$.
For the path estimator, the gradient is
\begin{equation}
\label{eq:grad-path}
\nabla_\theta \widehat{J}_{\textsc{Path}} = \sum_{k} \nabla_\theta W_k.
\end{equation}
For the branch estimator, the gradient has two parts:
\begin{equation}
\label{eq:grad}
\nabla_\theta \widehat{J}_{\textsc{Branch}} = \sum_{k} W_{k-1} \nabla_\theta a_k + \sum_{k} a_k \nabla_\theta W_{k-1}.
\end{equation}
Comparing the two gradients, the branch estimator also replaces the sampled $b_k$ with its conditional expectation $a_k$ in both terms, so the gradient variance of each prefix contribution is no larger than the path estimator's, as proved in Appendix~\ref{app:variance-proof}.

\subsection{The \real{} Training Loss}
\label{sec:method-real}

We define \real{}, our final training loss based on the branch estimator:
\begin{equation}
\label{eq:real-loss}
\mathcal{L}_{\real{}} =
\lambda \left(1 - \frac{\widehat{J}_{\textsc{Branch}}}{\gamma}\right)
+ (1-\lambda)\,\ell_{\mathrm{CE}}.
\end{equation}
The fixed window size $\gamma$ normalizes the \EAL{} term, and the CE term stabilizes gradients early in training (ablated in Section~\ref{sec:results-ablation}).
Implementing \real{} requires changing only the training loss, leaving the drafter architecture, training pipeline, and inference procedure unchanged, with no additional overhead.
Further implementation details are provided in Appendix~\ref{app:impl}.

\section{Experiments}
\label{sec:experiments}

\subsection{Experimental Setup}
\label{sec:setup}

\paragraph{Target and Draft Models.}
We use the Speculators framework \cite{speculators} for training, with the frozen Qwen3.6-35B-A3B target model \cite{qwen36_35b_a3b} and the \domino{} drafter \cite{huang2026domino} in mostly the same setting as the original \domino{} paper, changing only the loss and the draft block size to 8 (1 anchor and $\gamma = 7$).
We select this target because its mixture-of-experts (MoE) architecture and hybrid Gated Delta Network (GDN) structure reflect architectural choices in modern LLMs, with the aim of making our findings more transferable to current mainstream model architectures.
We select \domino{} as the drafter because it slightly outperformed other baselines in our preliminary comparisons.

\paragraph{Data and training.}
We use the Open-PerfectBlend prompts \cite{labonne2024openperfectblend,xu2024perfectblend} for training, with continuations regenerated by the target model at temperature 1.0.
We use Muon \cite{jordan2024muon} for hidden-layer weight matrices with learning rate $2\times10^{-4}$ and AdamW \cite{loshchilov2019decoupled} for the remaining trainable parameters with learning rate $1\times10^{-4}$.
All drafters are trained from scratch for 3 epochs (140K steps) with fixed $\lambda = 0.9$.

\paragraph{Evaluation protocol.}
We evaluate on 9 task subsets with temperature 1.0: HumanEval \cite{chen2021evaluating}, GSM8K \cite{cobbe2021training}, Natural Questions \cite{kwiatkowski2019natural}, RAG \cite{karpukhin2020dense}, CNN/DailyMail \cite{nallapati2016abstractive}, BFCL \cite{patil2025berkeley}, WMT (German--English) \cite{bojar2014findings}, MATH-500 \cite{lightman2024lets} and MBPP \cite{austin2021mbpp}.
The serving acceptance length measured by the vLLM engine \cite{kwon2023efficient} is $\mathrm{AL}=1+N_{\mathrm{accepted}}/N_{\mathrm{rounds}}$ (the $+1$ is the bonus token).
Each checkpoint is evaluated under two serving protocols: \emph{probabilistic} serving, which samples proposals and applies the $p/q$ rejection rule (our modeled setting), and \emph{greedy} serving, which proposes the draft argmax while the target still samples, with results averaged over 5 sampling seeds and 3 repetitions per seed (see Appendix~\ref{app:seeds} for the full protocol).

\paragraph{Compared methods.}
We compare \EAL{}-class objectives and the \slk{} baseline.
All methods use the same auxiliary CE component, and all baselines are reimplemented at the loss level:
\begin{itemize}[leftmargin=1.2em,itemsep=0pt]
\item \textbf{\slk{}}, a static variant of LK \cite{samarin2026lk}, with the same CE/TV components as the \dspark{} loss \cite{cheng2026dspark}.
\item \textbf{\wtv{}} \cite{xia2026acceptance}, a surrogate that multiplies per-step acceptance rates over the proposal window.
\item \textbf{\vsd{}} \cite{zou2026variational}, a variational lower bound on the acceptance rate over draft paths.
\item \textbf{\dblast{}} \cite{karimi2026dblast}, an acceptance-oriented loss over log importance ratios and log suffix acceptance length.
\item \textbf{\ealb{}}, our real \EAL{} training loss.
\item \textbf{\ealp{}}, a variant of \ealb{} without Rao--Blackwellization.
\end{itemize}
Appendix~\ref{app:setup} gives the full shared configuration and the loss recipe for each method.

\subsection{Main Results}
\label{sec:results}

\begin{table*}[t]
\centering
\setlength{\tabcolsep}{2pt}
\begin{tabular}{l c@{\hspace{4pt}} ccccccccc}
\toprule
Method & Avg & HumanEval & GSM8K & NQ & RAG & CNN/DM & BFCL & WMT & MATH-500 & MBPP \\
\midrule
\slk{} & 4.740 & 5.227 & 5.899 & 3.643 & 4.449 & 4.171 & 3.968 & 4.542 & 5.752 & 5.011 \\
\wtv{} & 4.715 & 5.208 & 5.891 & 3.610 & 4.426 & 4.144 & 3.923 & 4.522 & 5.744 & 4.967 \\
\vsd{} & 4.657 & 5.122 & 5.963 & 3.514 & 4.356 & 4.017 & 3.956 & 4.383 & 5.716 & 4.887 \\
\dblast{} & 4.749 & 5.222 & 5.998 & 3.615 & 4.437 & 4.155 & 4.022 & 4.511 & 5.789 & 4.992 \\
\ealp{} & \textbf{4.791} & 5.264 & \textbf{6.030} & \textbf{3.649} & 4.498 & 4.196 & \textbf{4.023} & 4.574 & \textbf{5.846} & 5.041 \\
\ealb{} & \textbf{4.792} & \textbf{5.270} & 6.021 & 3.645 & \textbf{4.499} & \textbf{4.218} & 4.007 & \textbf{4.590} & 5.831 & \textbf{5.047} \\
\bottomrule
\end{tabular}

\caption{Probabilistic acceptance length after 3 training epochs, averaged over 5 seeds.
Avg is the equal-weight mean over 9 task subsets.
\textbf{Bold} values indicate the top two results in the Avg column and the best result in each subset column.}
\label{tab:main}
\end{table*}

Table~\ref{tab:main} presents our main results, with overall and per-subset probabilistic acceptance length across 9 task subsets.
\ealb{} and its variant without RB top the table under probabilistic serving, reaching 4.792 and 4.791, respectively, ahead of all other compared methods (4.657--4.749).
Under greedy serving, \slk{} attains the highest AL at 4.544 (Table~\ref{tab:subset_greedy}), and among the \EAL{}-class objectives, \ealb{} performs best (4.537), while \ealp{} trails at 4.515.
The per-subset results under probabilistic serving (Table~\ref{tab:main}) show that on every subset, the best acceptance length is attained by either \ealb{} or \ealp{}.
\ealb{} and \ealp{} also rank in the top two for end-to-end serving throughput, with full results provided in Appendix~\ref{app:vllm-serving}.
This consistent leading performance across diverse tasks demonstrates the effectiveness of our real acceptance length objective.

\begin{figure}[t]
\centering
\includegraphics[width=0.83\columnwidth]{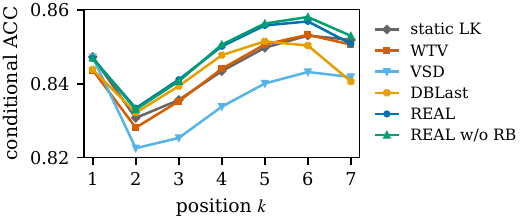}
\caption{Conditional acceptance $c_k$ at the three-epoch checkpoint, defined as the probability of accepting the $k$th draft token given the first $k-1$ tokens are accepted.}
\label{fig:retention-conditional}
\end{figure}

\paragraph{Position-wise Acceptance.}
As an auxiliary position-wise diagnostic, we analyze the 9-subset evaluation under probabilistic serving at $T=1.0$ (Table~\ref{tab:position-retention} and Figure~\ref{fig:retention-conditional}).
At $k=1$, \slk{}, \vsd{}, \ealb{}, and \ealp{} are nearly identical, with \slk{} slightly ahead.
\ealb{} leads at $k=2$--$4$, while \ealp{} leads at $k=5$--$7$.
At $k=7$, \ealp{} and \ealb{} reach $S_7=0.316$ and $0.315$, respectively, compared with $0.306$ for \slk{} and $0.304$ for \wtv{}.
Figure~\ref{fig:retention-conditional} shows the clear advantage of \ealb{} and \ealp{} in conditional acceptance, indicating stronger continuation of accepted prefixes.
The slight first-position edge of \slk{} is consistent with its single-step objective, while the stronger retention of \ealb{} and \ealp{} at later positions aligns with their exact \EAL{} objective over the proposal window.
This pattern indicates that the exact objective's advantage comes primarily from longer accepted prefixes rather than only the first position.

\begin{table}[t]
\centering
\setlength{\tabcolsep}{2pt}
\begin{tabular}{c cccccc}
\toprule
$k$ & \slk{} & \wtv{} & \vsd{} & \dblast{} & \ealb{} & \ealp{} \\
\midrule
1 & \textbf{0.847} & 0.844 & 0.847 & 0.844 & 0.847 & 0.847 \\
2 & 0.704 & 0.699 & 0.697 & 0.702 & \textbf{0.706} & 0.705 \\
3 & 0.588 & 0.584 & 0.575 & 0.589 & \textbf{0.594} & 0.593 \\
4 & 0.496 & 0.493 & 0.480 & 0.500 & \textbf{0.505} & 0.504 \\
5 & 0.421 & 0.419 & 0.403 & 0.425 & 0.432 & \textbf{0.432} \\
6 & 0.360 & 0.357 & 0.340 & 0.362 & 0.370 & \textbf{0.371} \\
7 & 0.306 & 0.304 & 0.286 & 0.304 & 0.315 & \textbf{0.316} \\
\bottomrule
\end{tabular}

\caption{Acceptance retention $S_k$ at the three-epoch checkpoint, defined as the probability of accepting at least $k$ draft tokens in a speculative round.
\textbf{Bold} values mark the highest unrounded retention at each position.}
\label{tab:position-retention}
\end{table}

\subsection{Training Diagnostics}
\label{sec:results-diagnostics}

\begin{figure}[t]
\centering
\includegraphics[width=0.6\columnwidth]{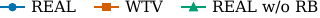}\par
\vspace{2pt}
\subcaptionbox{training \EAL{} proxy\label{fig:training-eal}}{\includegraphics[width=0.494\columnwidth]{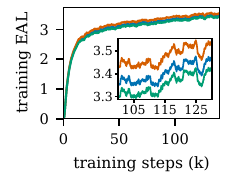}}
\subcaptionbox{serving AL\label{fig:serving-epoch}}{\includegraphics[width=0.494\columnwidth]{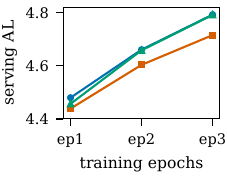}}\hfill
\caption{Training \EAL{} proxy and serving AL during training for \wtv{}, \ealb{}, and \ealp{}.
\ealb{} leads on probabilistic serving AL, while \wtv{} leads on the training \EAL{} proxy.}
\label{fig:al-training}
\end{figure}

\begin{figure}[t]
\centering
\subcaptionbox{\ealb{}\label{fig:jb_ealb}}{\includegraphics[width=0.33\columnwidth]{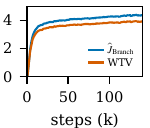}}\hfill
\subcaptionbox{\wtv{}\label{fig:jb_wtv}}{\includegraphics[width=0.33\columnwidth]{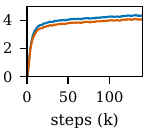}}\hfill
\subcaptionbox{\ealp{}\label{fig:jb_eala}}{\includegraphics[width=0.33\columnwidth]{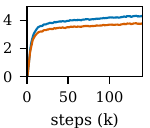}}
\caption{The branch \EAL{} estimate $\widehat{J}_{\textsc{Branch}}$ and the \wtv{} surrogate during training for \wtv{}, \ealb{}, and \ealp{}.
The \wtv{} surrogate stays below the branch estimate in all three cases.}
\label{fig:jb_jwtv}
\end{figure}

\begin{figure}[t]
\centering
\includegraphics[width=0.7\columnwidth]{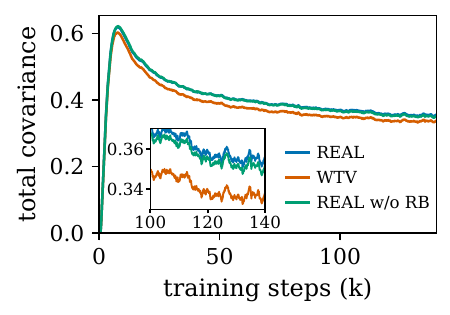}
\caption{Total covariance summed over all 6 adjacent pairs during training for \wtv{}, \ealb{}, and \ealp{}.
\ealb{} and \ealp{} track each other closely and stay above \wtv{} throughout training.}
\label{fig:cov}
\end{figure}

\paragraph{Acceptance Length.}
Figure~\ref{fig:al-training} compares the argmax-based training \EAL{} proxy (Figure~\ref{fig:training-eal}), defined as $\sum_k \prod_{i \leq k} \mathrm{acc}_i$ where $\mathrm{acc}_i$ is the argmax accuracy at position $i$, with serving acceptance length (AL) at the end of each epoch (Figure~\ref{fig:serving-epoch}).
The training \EAL{} proxy favors \wtv{} throughout training, which is expected because this metric shares the product-of-marginals form of the \wtv{} objective, while \ealb{} leads \wtv{} on probabilistic serving AL.
This opposite ordering between the training \EAL{} proxy and serving AL shows that the \wtv{}-form metric is not reliable for judging draft-training objectives.
\ealp{} reads slightly below \ealb{} on the training \EAL{} proxy and at epoch 1 in serving AL, but the two nearly coincide from epoch 2 onward, suggesting that Rao--Blackwellization helps mainly early in training.
We also track the branch \EAL{} estimate $\widehat{J}_{\textsc{Branch}}$ and the \wtv{} surrogate during training, finding that \wtv{} consistently reads below the branch estimate for all three methods (Figure~\ref{fig:jb_jwtv}).
This shows that our exact objective recovers acceptance length that the surrogate misses, consistent with the positive covariances observed throughout training (Figure~\ref{fig:cov}).

\paragraph{Covariance Analysis.}

Motivated by equation~\eqref{eq:cov}, we log adjacent-pair covariances on training batches to examine whether path weights and continuation acceptance are associated during training.
Figure~\ref{fig:cov} shows the total covariance summed over all 6 adjacent pairs for \wtv{}, \ealb{}, and \ealp{} during training, each with an early spike, a decline, and a non-zero plateau.
All methods maintain positive covariances throughout training, with \ealb{} and \ealp{} nearly identical and both consistently above \wtv{}.
This is consistent with the gap in Figure~\ref{fig:jb_jwtv}, which confirms that the \wtv{} surrogate systematically underestimates the exact objective.
Detailed per-pair analysis is provided in Appendix~\ref{app:cov-pairs}.

\subsection{Ablations}
\label{sec:results-ablation}

We ablate four aspects: Rao--Blackwellization, the auxiliary CE term, serving temperature, and the training budget.

\paragraph{Rao--Blackwellization.}
\ealp{} ablates Rao--Blackwellization, keeping the sampled $b_k$ in place of the marginalized $a_k$.
As shown in Table~\ref{tab:rb-variance-main}, the per-term gradient variance is lower at every position in every checkpoint, with reductions of 76--84\% at $k=2$ and 52--58\% for the aggregate gradient.
Rao--Blackwellization preserves the expected gradient while guaranteeing no greater covariance than the path estimator for each prefix-gradient contribution (Section~\ref{sec:method-ealb}).
As shown in Tables~\ref{tab:main} and~\ref{tab:subset_greedy}, \ealp{} reaches 4.791 probabilistic AL, comparable to \ealb{} (4.792), while its greedy AL (4.515) trails \ealb{} (4.537).
Despite their similar final probabilistic AL at $T=1$, \ealb{} performs better early in training (Figure~\ref{fig:al-training}) and generalizes better across serving temperatures (Table~\ref{tab:temp-sweep}).
The pilot study further illustrates the optimization difficulties of sampled-path estimation (Appendix~\ref{app:toy}).

\begin{table}[t]
\centering
\setlength{\tabcolsep}{3pt}
\begin{tabular}{l ccccccc c}
\toprule
epoch & $k{=}1$ & $k{=}2$ & $k{=}3$ & $k{=}4$ & $k{=}5$ & $k{=}6$ & $k{=}7$ & agg \\
\midrule
\multicolumn{9}{l}{checkpoints of \ealb{}} \\
ep1 & 0 & 0.17 & 0.44 & 0.45 & 0.61 & 0.56 & 0.73 & 0.47 \\
ep2 & 0 & 0.17 & 0.38 & 0.49 & 0.59 & 0.65 & 0.66 & 0.48 \\
ep3 & 0 & 0.20 & 0.40 & 0.48 & 0.47 & 0.55 & 0.67 & 0.43 \\
\midrule
\multicolumn{9}{l}{checkpoints of \ealp{}} \\
ep1 & 0 & 0.24 & 0.44 & 0.45 & 0.42 & 0.58 & 0.68 & 0.47 \\
ep2 & 0 & 0.16 & 0.35 & 0.45 & 0.61 & 0.48 & 0.64 & 0.42 \\
ep3 & 0 & 0.19 & 0.36 & 0.46 & 0.54 & 0.63 & 0.62 & 0.44 \\
\bottomrule
\end{tabular}

\caption{$\mathrm{Var}_{\textsc{Branch}}/\mathrm{Var}_{\textsc{Path}}$ for per-term gradient variance at each position and for aggregate gradient variance.
Lower values indicate greater variance reduction.
The zero at $k{=}1$ is structural: given the fixed prefix, $a_1$ is deterministic while $b_1$ is sampled.}
\label{tab:rb-variance-main}
\end{table}

\begin{figure}[t]
\centering
\includegraphics[width=0.6\columnwidth]{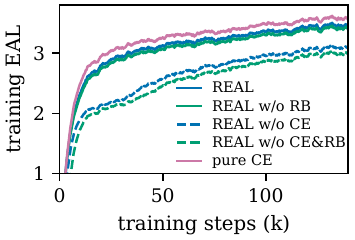}
\caption{Training \EAL{} proxy trajectories of \ealb{} and \ealp{} with and without the auxiliary CE term, and of pure CE.}
\label{fig:ce-ablation}
\end{figure}

\paragraph{Auxiliary CE term.}
To isolate whether the serving gains come from the \EAL{} term or merely from the auxiliary CE term, we train pure CE without the \EAL{} term, as well as \ealb{} w/o CE and \real{} w/o CE\&RB.
Both variants without CE train to completion but stay below their CE-augmented counterparts in greedy and probabilistic serving AL (Table~\ref{tab:ce-ablation}), as well as in the training \EAL{} proxy trajectory (Figure~\ref{fig:ce-ablation}), showing that the auxiliary CE term stabilizes training and improves serving performance with and without RB.
Although pure CE outperforms \ealb{} on the training \EAL{} proxy (Figure~\ref{fig:ce-ablation}), its greedy and probabilistic serving AL are again below both \ealb{} and \ealp{} (Table~\ref{tab:ce-ablation}), demonstrating the effectiveness of the \EAL{} loss term.
Thus, both the \EAL{} term and the auxiliary CE term contribute to the final performance.

\begin{table}[t]
\centering
\begin{tabular}{l cc}
\toprule
 & \textbf{greedy} & \textbf{probabilistic} \\
\midrule
\real{} w/o CE\&RB & 4.369 & 4.588 \\
\real{} w/o CE & 4.397 & 4.624 \\
pure CE & 4.508 & 4.668 \\
\ealp{} & 4.515 & 4.791 \\
\ealb{} & \textbf{4.537} & \textbf{4.792} \\
\bottomrule
\end{tabular}

\caption{Ablation of the auxiliary CE term at the three-epoch budget.
Removing CE from \ealb{} or \ealp{}, or using pure CE without the \EAL{} term, degrades serving AL under both protocols.}
\label{tab:ce-ablation}
\end{table}

\paragraph{Serving temperature.}
We evaluate the compared methods at $T \in \{1.0, 0.7, 0.5\}$ for both the draft and target models to examine how serving temperature affects the advantage of our method (Table~\ref{tab:temp-sweep}).
As the temperature decreases, all methods achieve higher acceptance length, while \ealb{}'s advantage over the baselines narrows and \slk{} surpasses it at $T = 0.5$.
Meanwhile, \ealp{} is more sensitive to temperature change, falling further behind \ealb{} as the temperature decreases.
This trend matches our expectation that the main advantage of \real{} lies in stochastic serving, because it is derived for stochastic rejection sampling and its benefit diminishes as the serving temperature decreases and decoding approaches the fully deterministic regime, where \slk{} is strongest.
\begin{table}[t]
\centering
\begin{tabular}{l ccc}
\toprule
method & $T = 1.0$ & $T = 0.7$ & $T = 0.5$ \\
\midrule
\slk{} & 4.740 & 4.925 & \textbf{5.002} \\
\wtv{} & 4.715 & 4.867 & 4.981 \\
\vsd{} & 4.657 & 4.842 & 4.928 \\
\dblast{} & 4.749 & 4.912 & 4.975 \\
\ealp{} & 4.791 & 4.919 & 4.964 \\
\ealb{} & \textbf{4.792} & \textbf{4.935} & 4.993 \\
\bottomrule
\end{tabular}

\caption{Probabilistic acceptance length across serving temperatures at the three-epoch budget, averaged over the 5 seeds and 9 subsets.
\textbf{Bold} values mark the best method per temperature.}
\label{tab:temp-sweep}
\end{table}

\begin{figure}[t]
\centering
\includegraphics[width=0.7\columnwidth]{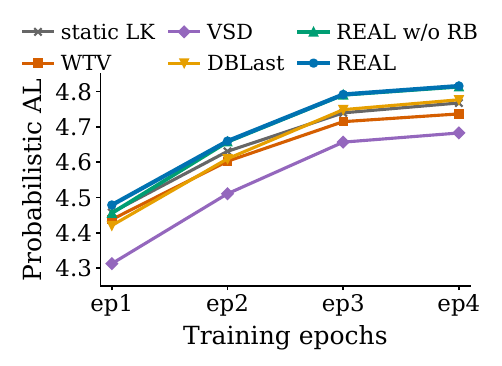}
\caption{Probabilistic acceptance length at the end of each epoch, including a continuation to epoch 4, averaged over 9 subsets and 5 seeds.
\ealb{} attains the highest AL at every epoch.}
\label{fig:epoch}
\end{figure}

\paragraph{Training budget.}
To check whether our \EAL{} objective affects convergence speed and sustained gains, we measure probabilistic serving AL at the end of each epoch and continue training for one additional epoch, as shown in Figure~\ref{fig:epoch}.
The full results under both protocols are provided in Appendix~\ref{app:epoch}.
\ealb{} attains the highest probabilistic AL at every epoch, and its lead over \wtv{} grows from $0.041$ at epoch 1 to $0.077$ at epoch 3 (4.479 vs. 4.438 and 4.792 vs. 4.715, respectively).
Both \ealb{} and \ealp{} maintain their lead through the additional epoch and improve on both protocols, reaching 4.816 and 4.814 probabilistic AL, respectively (Appendix~\ref{app:epoch}), so the gains of the exact objective accumulate throughout training rather than saturate early.

\section{Related Work}
\label{sec:related_work}

\subsection{Draft Architectures}
\label{sec:draft_arch}

Draft designs include independent heads \cite{cai2024medusa}, sequentially dependent heads \cite{ankner2024hydra}, feature-level autoregression \cite{li2024eagle,li2024eagle2,li2025eagle3}, and low-overhead variants \cite{du2024glide}.
Block-parallel drafters generate a block of tokens per round, evolving from early blockwise decoding \cite{stern2018blockwise} and multi-token prediction heads \cite{gloeckle2024better} to \dflash{} \cite{chen2026dflash}, \domino{} \cite{huang2026domino}, \dspark{} \cite{cheng2026dspark}, and P-EAGLE \cite{hui2026peagle}, with diffusion-based variants enabling bidirectional generation \cite{li2025diffuspec,sandler2025specdiff2,whalen2026teaching,liu2026angelspec}.

\subsection{Training Objectives of Draft Models}
\label{sec:train_objectives}

Draft models are commonly trained with cross-entropy or forward KL distillation on teacher-forced contexts \cite{zhou2023distillspec}, with on-policy variants using draft rollouts \cite{lei2026draftopd,liu2023online}.
These distribution-matching objectives optimize token-level similarity rather than acceptance length.

Single-step total variation objectives instead maximize the one-step acceptance rate in Equation~\eqref{eq:tv} \cite{goel2024direct,samarin2026lk}.
Multi-step variants, including end-to-end TV \cite{li2026bebop}, DFly's TV stage \cite{liu2026angelspec}, and WTV \cite{xia2026acceptance}, multiply per-step acceptance rates without preserving the accepted-token distribution in Equation~\eqref{eq:r}.

Other acceptance-oriented methods reweight positions in cross-entropy \cite{wu2026dpace,gu2026verification}, use acceptance-oriented or variational surrogates \cite{karimi2026dblast,zou2026variational}, or adopt target-aligned training and diffusion-based rewards \cite{an2026pard2,sandler2025specdiff2,whalen2026teaching}.

Concurrent work includes EDR, which minimizes exact expected decoding rounds using additional dynamic programming during training \cite{zhao2026edr}, BV loss, which targets acceptance under block verification \cite{kim2026bvloss}, and FReDraft, which uses draft-generated rollouts with acceptance marginalization and an over-sampled-token penalty \cite{anonymous2026fredraft}.
\real{} optimizes \EAL{} under standard token verification by changing only the loss in the existing training pipeline, without being tied to a specific rollout or verification method or incurring additional training overhead.

\subsection{Other Speculative Decoding Optimization Methods}
\label{sec:other_methods}

Other work optimizes serving through draft-target separation \cite{shi2024greenllm,park2025specedge,zhang2025swiftspec,kumar2026ssd}, verification and speculation control \cite{suresh2024blockverif,huang2024specdecpp,liu2024turbospec}, or tree-structured proposals \cite{miao2023specinfer,chen2024sequoia,li2024eagle2,cai2024medusa,wang2025opttree,hu2026gto}, which are orthogonal to our focus on the training objective.

\section{Conclusion}
\label{sec:conclusion}

We have presented \real{}, a real \EAL{} training loss based on an unbiased and differentiable branch estimator for stochastic rejection sampling, using path weights to restore the propose-and-accept branch structure and Rao--Blackwellization to reduce per-term variance.
Our covariance analysis explains why products of marginal acceptance rates can reverse the optimization direction.
With only a change to training loss and no additional overhead, \ealb{} achieves the best probabilistic acceptance length among the main compared objectives across 9 task subsets and 5 seeds at $T=1$, demonstrating its effectiveness.

\section*{Limitations}
\label{sec:limitations}

The evaluation is limited to one target model, one drafter architecture, and one sampled data recipe to enable controlled comparisons; broader evaluation across these settings is left for future work.
The \vsd{} and DBLast comparisons use loss-level adaptations rather than full system reproductions, isolating the effect of the loss itself.
Our objective primarily targets probabilistic serving, and \ealb{} is near-optimal under greedy serving.

\section*{Acknowledgments}

The author thanks Jiale Fu for insightful discussions, help with checking the mathematical derivations, and valuable suggestions on the organization and presentation of this work.

\bibliography{references}

\appendix
\section{Counterexample Details}
\label{app:counterexample}

This appendix gives the full two-step enumeration behind the counterexample of Section~\ref{sec:method-gap} and a second example in which the surrogate reverses the ranking of two drafts.

\subsection{Full Enumeration of the Two-Step Counterexample}

The setup of Section~\ref{sec:method-gap}: vocabulary $\{A, B\}$, $p_1 = (0.4, 0.6)$, $q_1 = (0.8, 0.2)$, $p_2(\cdot \mid hA) = (0.9, 0.1)$, $p_2(\cdot \mid hB) = (0.1, 0.9)$, and $q_2 = (\theta, 1-\theta)$ on both branches with $0.1 < \theta < 0.9$.
The per-token quantities of step 1 are:

\begin{center}
\begin{tabular}{l ccc}
\toprule
token & $p_1$ & $q_1$ & $m_1 = \min(p_1, q_1)$ \\
\midrule
$A$ & 0.4 & 0.8 & 0.4 \\
$B$ & 0.6 & 0.2 & 0.2 \\
\bottomrule
\end{tabular}
\end{center}

\noindent so $a_1 = 0.6$, and the path weights are $b_1(A) = \min(1, 0.8/0.4) = 1$ and $b_1(B) = \min(1, 0.2/0.6) = 1/3$.
The second-step acceptance rates are $a_2(hA) = \min(0.9, \theta) + \min(0.1, 1-\theta) = \theta + 0.1$ and $a_2(hB) = \min(0.1, \theta) + \min(0.9, 1-\theta) = 1.1 - \theta$.
Enumerating all four two-token paths:

\begin{center}
\begin{tabular}{l cc}
\toprule
path & target probability & propose--accept mass \\
\midrule
$AA$ & $0.4 \times 0.9 = 0.36$ & $0.4\theta$ \\
$AB$ & $0.4 \times 0.1 = 0.04$ & $0.04$ \\
$BA$ & $0.6 \times 0.1 = 0.06$ & $0.2 \times 0.1 = 0.02$ \\
$BB$ & $0.6 \times 0.9 = 0.54$ & $0.2(1-\theta)$ \\
\bottomrule
\end{tabular}
\end{center}

\noindent Hence $P(L \ge 2) = 0.26 + 0.20\theta$ and $J_2 = 0.86 + 0.20\theta$.
The surrogate averages the second step over target proportions: $\widetilde{J}_{\mathrm{\wtv{}},2} = 0.6 + 0.24\,a_2(hA) + 0.36\,a_2(hB) = 1.02 - 0.12\theta$.
The branch estimator on target samples reads $\widehat{J}(A) = 0.6 + 1 \times (0.1 + \theta)$ and $\widehat{J}(B) = 0.6 + \tfrac{1}{3}(1.1 - \theta)$, whose target average $0.4\widehat{J}(A) + 0.6\widehat{J}(B)$ equals $J_2$ exactly.

\paragraph{Gradient variance in the same example.}
At $\theta = 0.5$, the per-path gradients give $\mathbb{E}[g_{\textsc{Branch}}] = +0.20$ with variance $32/75 \approx 0.427$, $\mathbb{E}[g_{\mathrm{\wtv{}}}] = -0.12$ with variance $0.3456$, and $\mathbb{E}[g_{\textsc{Path}}] = +0.20$ with variance $323/675 \approx 0.479$.
The surrogate has the lowest gradient variance but a biased mean: with $M$ independent trajectories its gradient MSE is $64/625 + 216/(625M)$, which does not vanish as $M$ grows, while the exact estimators' MSE vanishes as $1/M$.
Lower variance does not compensate for a wrong mean.

\subsection{The Surrogate Can Reverse Model Rankings}
\label{app:ranking}

Take a first step with target $p_1 = (0.5, 0.5)$ and two candidate drafts $q_1^{(1)} = (0.70, 0.30)$ and $q_1^{(2)} = (0.35, 0.65)$.
The second step has $p_2(\cdot \mid hA) = (0.5, 0.5)$ with $q_2(\cdot \mid hA) = (0.6, 0.4)$, and $p_2(\cdot \mid hB) = (0.05, 0.95)$ with $q_2(\cdot \mid hB) = (0.95, 0.05)$, so $a_2(hA) = 0.9$ and $a_2(hB) = 0.1$.
Draft 1 has the lower single-step acceptance rate ($a_1^{(1)} = 0.8 < a_1^{(2)} = 0.85$) but the higher true objective:
\begin{center}
\begin{tabular}{l cc}
\toprule
 & Draft 1 & Draft 2 \\
\midrule
$a_1$ & 0.80 & 0.85 \\
$\widetilde{J}_{\mathrm{\wtv{}},2}$ & 1.200 & \textbf{1.275} \\
$J_2$ (exact) & \textbf{1.280} & 1.215 \\
\bottomrule
\end{tabular}
\end{center}
The surrogate ranks Draft 2 above Draft 1; the exact objective ranks them oppositely.
The bias is not a shared constant offset: it changes which model is better.

\section{Proof of the Variance Guarantee}
\label{app:variance-proof}

We provide a detailed proof of the variance guarantee stated in Section~\ref{sec:method-ealb}.

\paragraph{Setup.}
For a fixed prefix $h$, define the $k$-th prefix contribution of the path estimator and the branch estimator as
\begin{equation}
T_k^{\textsc{Path}} = W_{k-1} b_k, \quad T_k^{\textsc{Branch}} = W_{k-1} a_k,
\end{equation}
where $W_{k-1} = \prod_{i=1}^{k-1} b_i$ and $a_k = \mathbb{E}[b_k \mid h_k]$.
Both estimators are unbiased: $\mathbb{E}[T_k^{\textsc{Path}}] = \mathbb{E}[T_k^{\textsc{Branch}}] = S_k$, where $S_k = P(L \ge k)$ is the survival probability.

\phantomsection
\paragraph{Gradient variance reduction for both terms.}
\label{prop:grad-variance-both}
The gradient of the path estimator is
\begin{equation}
g_k^{\textsc{Path}} = b_k \nabla_\theta W_{k-1} + W_{k-1} \nabla_\theta b_k,
\end{equation}
and the gradient of the branch estimator is
\begin{equation}
g_k^{\textsc{Branch}} = a_k \nabla_\theta W_{k-1} + W_{k-1} \nabla_\theta a_k.
\end{equation}
For either estimator $s\in\{\textsc{Path},\textsc{Branch}\}$, define its per-term gradient variance as
\begin{equation}
\mathrm{Var}_k^s = \mathbb{E}\!\left[\left\|g_k^s-\mathbb{E}[g_k^s]\right\|^2\right].
\end{equation}
Define the gradient residual
\begin{equation}
r_k = g_k^{\textsc{Path}} - g_k^{\textsc{Branch}}.
\end{equation}
The branch gradient is the conditional mean $g_k^{\textsc{Branch}}=\mathbb{E}[g_k^{\textsc{Path}}\mid h_k]$, so $\mathbb{E}[r_k \mid h_k] = 0$.
Let $\mu_k=\mathbb{E}[g_k^{\textsc{Branch}}]=\mathbb{E}[g_k^{\textsc{Path}}]$ denote their common mean.
We write the gradient error as
\begin{equation}
g_k^{\textsc{Path}} - \mu_k = (g_k^{\textsc{Branch}} - \mu_k) + r_k.
\end{equation}
Taking the expectation of the squared norm gives
\begin{equation}
\mathrm{Var}_k^{\textsc{Path}} = \mathrm{Var}_k^{\textsc{Branch}} + \mathbb{E}\!\left[\|r_k\|^2\right].
\end{equation}
Since $g_k^{\textsc{Branch}}-\mu_k$ is fixed given $h_k$, the cross term vanishes after taking expectation:
\begin{equation}
\begin{aligned}
&\mathbb{E}\!\left[\left\langle g_k^{\textsc{Branch}}-\mu_k,r_k\right\rangle\right]\\
&\quad=\mathbb{E}\!\left[\left\langle g_k^{\textsc{Branch}}-\mu_k,\mathbb{E}[r_k\mid h_k]\right\rangle\right]=0.
\end{aligned}
\end{equation}
Therefore, $\mathrm{Var}_k^{\textsc{Branch}}\leq \mathrm{Var}_k^{\textsc{Path}}$ for every prefix term $k$.

\section{Per-Pair Covariance Trajectories}
\label{app:cov-pairs}

\begin{figure}[t]
\centering
\includegraphics[width=0.95\columnwidth]{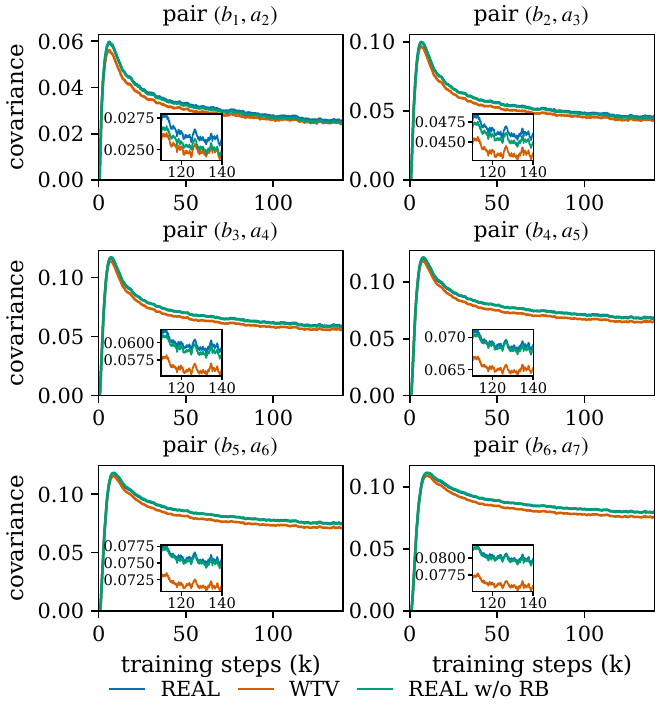}
\caption{Per-pair covariance trajectories during training for \wtv{}, \ealb{}, and \ealp{}.
Each panel shows one adjacent pair $(b_k, a_{k+1})$, $k = 1, \dots, 6$.}
\label{fig:cov-pairs}
\end{figure}

This appendix shows the per-pair covariance trajectories behind the summed diagnostic of Figure~\ref{fig:cov}, one panel per adjacent pair (Figure~\ref{fig:cov-pairs}).
Each position follows the same overall trend as the summed diagnostic, with generally larger covariances at later positions.

\section{Implementation Details}
\label{app:impl}

This appendix collects the implementation details: the shared setup and loss recipes, ratio direction, normalization and masks, the logging diagnostics, and the verifier consistency check.

\subsection{Shared Setup and Loss Recipes}
\label{app:setup}

The main-study methods share the training data, drafter architecture, optimizer, and training budget (Table~\ref{tab:setup}), and differ in the training loss (Table~\ref{tab:recipes}).
The shared base-anchored curriculum evaluates each method's loss recipe on both the pre-bias logits $z_{\mathrm{base}}$ and the final GRU-refined logits $z_{\mathrm{final}}$:
\begin{equation}
\begin{aligned}
\mathcal{L}_t ={}&(1-c_t)\mathcal{L}_{\mathrm{recipe}}(z_{\mathrm{final}}) \\
&+ c_t\mathcal{L}_{\mathrm{recipe}}(z_{\mathrm{base}}),
\end{aligned}
\end{equation}
where $c_t = \lambda_{\mathrm{base}}$ decreases linearly from 1 to 0 over the main training budget.
This coefficient is separate from the fixed mixture coefficient $\lambda$ within the \ealb{} recipe.
The variants without CE remove the auxiliary CE component from the recipe on both branches, while keeping the same curriculum.

\begin{table}[h]
\centering
\begin{tabular}{l p{0.64\columnwidth}}
\toprule
component & setting \\
\midrule
target model & frozen Qwen3.6-35B-A3B (MoE) \\
drafter & \domino{} block-parallel drafter, 5 layers on target layers 2/20/37 \\
 & GRU hidden dim 1024, embedding dim 256 \\
block & 8 tokens per round (1 anchor + $\gamma = 7$ proposals), 4096 anchors \\
training data & Open-PerfectBlend continuations regenerated by the target at $T = 1.0$ \\
optimizer & Muon for hidden-layer weight matrices, LR $2 \times 10^{-4}$ \\
 & AdamW for remaining trainable parameters, LR $1 \times 10^{-4}$ \\
training budget & 3 epochs ($\approx$140K steps) \\
batch & 16384 tokens per rank per step \\
curriculum & same loss recipe on base and final logits, $\lambda_{\mathrm{base}}: 1 \to 0$ over training \\
evaluation & 9 subsets, 80 requests each, max 2048 output tokens \\
sampling & $T = 1.0$, top-$p$ 0.95, top-$k$ 20 \\
proposals & greedy (argmax) and probabilistic ($p/q$ rejection) \\
\bottomrule
\end{tabular}

\caption{Shared training and evaluation setup for the main study.}
\label{tab:setup}
\end{table}

\begin{table}[h]
\centering
\setlength{\tabcolsep}{2pt}
\begin{tabular}{l p{0.56\columnwidth}}
\toprule
method & loss configuration \\
\midrule
\slk{} & $\{\mathrm{TV}: 0.9, \mathrm{CE}: 0.1\}$, the static LK objective used in the \dspark{} recipe \\
\wtv{} & $\{\wtv{}: 0.9, \mathrm{CE}: 0.1\}$ \\
\vsd{} & $\{\mathrm{VSD}: 0.9, \mathrm{CE}: 0.1\}$, weighted-likelihood M-step with detached path-acceptance weights, self-normalized across valid blocks and divided by $\gamma$ \\
\dblast{} & $\{\dblast{}: 0.9, \mathrm{CE}: 0.1\}$, acceptance threshold $\tau = 0.1$ \\
\ealp{} & $\{\widehat{J}_{\textsc{Path}}: 0.9, \mathrm{CE}: 0.1\}$ \\
\ealb{} & $\{\widehat{J}_{\textsc{Branch}}: 0.9, \mathrm{CE}: 0.1\}$ \\
\midrule
\real{} w/o CE\&RB & $\{\widehat{J}_{\textsc{Path}}: 1.0\}$ \\
\real{} w/o CE & $\{\widehat{J}_{\textsc{Branch}}: 1.0\}$ \\
pure CE & $\{\mathrm{CE}: 1.0\}$ \\
\bottomrule
\end{tabular}

\caption{Loss configurations of the final experiments.
Each $\widehat{J}$ entry denotes the corresponding loss $1-\widehat{J}/\gamma$.}
\label{tab:recipes}
\end{table}

\paragraph{Ratio direction.}
The training-time path weight is $b_i = \min(1, q(y_i)/p(y_i))$ on target-sampled tokens; the inference-time acceptance probability is $\min(1, p(v)/q(v))$ on draft-proposed tokens.
The two ratios point in opposite directions and must not be interchanged.
Target log-probabilities are detached; draft log-probabilities are not.

\paragraph{Normalization and masks.}
The loss $1 - \widehat{J}/\gamma$ always divides by the fixed $\gamma$, never by the random effective length.
EOS counts as an acceptable token, positions after EOS are masked, and masks are prefix-contiguous so that the full-vocabulary sum in $a_k$ remains exact.
The prefix weight $W$ and the overlap $a$ both backpropagate (Equation~\eqref{eq:grad}).

\paragraph{Diagnostics.}
For each block, we log the branch \EAL{} estimate $\widehat{J}_{\textsc{Branch}}$ as \texttt{eal\_j\_b}, the \wtv{} surrogate value \texttt{eal\_j\_wtv} on the same trajectories, per-position means \texttt{eal\_a\_pos\_*} and \texttt{eal\_b\_pos\_*}, and per-pair \texttt{eal\_ba\_next\_pos\_*} for the covariance of Equation~\eqref{eq:cov}.
All diagnostic tensors are logged as sum/total pairs.

\paragraph{Verifier consistency.}
We evaluate two proposal strategies: argmax proposals (greedy) and sampled proposals with $p/q$ rejection (probabilistic).
Under probabilistic serving, a draft token $y \sim q$ is accepted with probability $\min(1, p(y)/q(y))$ \cite{leviathan2023fast}.
On rejection, the correction token is sampled from the normalized positive part of $p-q$.
Under greedy serving, the draft proposes its argmax token, but the target continues to sample from $p$.

\section{Recorded Training Cost}
\label{app:training-cost}

Table~\ref{tab:training-cost} summarizes the logged step times of the five EAL-class training objectives over three 10{,}000-step windows, one within each epoch, without outlier filtering.
The recorded step times are similar across these methods.
The timer covers batch fetch through optimizer completion, including synchronization and the shared objective diagnostics.
These measurements describe the recorded runs and do not isolate the incremental cost of the loss, since software builds and resource contention were not controlled for a timing comparison.

\begin{table}[t]
\centering
\setlength{\tabcolsep}{2pt}
\begin{tabular}{l l c c}
\toprule
method & global steps & median & IQR \\
\midrule
\wtv{} & 1{,}000--10{,}999 & 847.4 & [843.0, 855.4] \\
 & 50{,}000--59{,}999 & 850.2 & [845.2, 859.7] \\
 & 100{,}000--109{,}999 & 849.7 & [844.7, 859.1] \\
\midrule
\vsd{} & 1{,}000--10{,}999 & 834.7 & [830.0, 845.1] \\
 & 50{,}000--59{,}999 & 840.8 & [834.5, 854.2] \\
 & 100{,}000--109{,}999 & 845.9 & [838.5, 862.6] \\
\midrule
\dblast{} & 1{,}000--10{,}999 & 838.3 & [833.8, 847.5] \\
 & 50{,}000--59{,}999 & 839.5 & [834.9, 849.0] \\
 & 100{,}000--109{,}999 & 839.4 & [834.7, 849.7] \\
\midrule
\ealp{} & 1{,}000--10{,}999 & 835.5 & [831.4, 844.2] \\
 & 50{,}000--59{,}999 & 836.2 & [832.0, 845.7] \\
 & 100{,}000--109{,}999 & 838.6 & [834.0, 849.6] \\
\midrule
\ealb{} & 1{,}000--10{,}999 & 844.5 & [839.6, 854.2] \\
 & 50{,}000--59{,}999 & 846.7 & [841.2, 858.5] \\
 & 100{,}000--109{,}999 & 845.0 & [840.2, 855.2] \\
\bottomrule
\end{tabular}

\caption{Recorded training step time in milliseconds over three 10{,}000-step windows.
The bracketed interquartile range (IQR) $[Q_1,Q_3]$ spans the middle 50\% of step times.}
\label{tab:training-cost}
\end{table}

\section{Evaluation Seeds and Repetitions}
\label{app:seeds}

All checkpoints are evaluated with 5 sampling seeds $\{0, 777, 13, 1234, 999983\}$, with 3 repetitions per seed under probabilistic serving (15 evaluations per checkpoint) and 3 repetitions with the default seed under greedy serving.
Each repetition runs 80 requests per subset with at most 2048 output tokens, using the checkpoint's generation configuration ($T = 1.0$, top-$k$ 20, top-$p$ 0.95).
Reported values are equal-weight means over the subsets and seeds.

\begin{table}[t]
\centering
\setlength{\tabcolsep}{2pt}
\begin{tabular}{l ccccc c}
\toprule
method & 0 & 777 & 13 & 1234 & 999983 & mean \\
\midrule
\slk{} & 4.739 & 4.735 & 4.739 & 4.746 & 4.743 & 4.740 \\
\wtv{} & 4.714 & 4.726 & 4.706 & 4.715 & 4.714 & 4.715 \\
\vsd{} & 4.652 & 4.669 & 4.649 & 4.659 & 4.657 & 4.657 \\
\dblast{} & 4.757 & 4.748 & 4.745 & 4.744 & 4.751 & 4.749 \\
\ealp{} & 4.787 & 4.787 & 4.786 & \textbf{4.799} & \textbf{4.796} & 4.791 \\
\ealb{} & \textbf{4.804} & \textbf{4.788} & \textbf{4.787} & 4.797 & 4.783 & \textbf{4.792} \\
\bottomrule
\end{tabular}

\caption{Per-seed and seed-averaged probabilistic AL after three training epochs.
All methods are combined with the auxiliary CE term.
\textbf{Bold} marks the best value in each column.}
\label{tab:seed-stability}
\end{table}

Table~\ref{tab:seed-stability} shows the per-seed results.
The spread across seeds is at most 0.021 AL for every method, and \ealb{} and \ealp{} stay ahead of all other methods on every seed, so the main ranking is not driven by any particular seed.

\section{End-to-End Serving Throughput}
\label{app:vllm-serving}

Table~\ref{tab:vllm-serving} reports aggregate output throughput with a batch size of 80 for the same checkpoints and protocol (probabilistic serving, 5 seeds, 9 subsets, 3 repetitions each), with \ealp{} and \ealb{} taking the top two spots and supporting acceptance length as a speedup proxy.

\begin{table}[t]
\centering
\begin{tabular}{l c}
\toprule
method & Throughput (tokens/s) \\
\midrule
\slk{} & 4622.1 \\
\wtv{} & 4553.0 \\
\vsd{} & 4535.8 \\
\dblast{} & 4648.6 \\
\ealp{} & \textbf{4664.8} \\
\ealb{} & \textbf{4662.0} \\
\bottomrule
\end{tabular}

\caption{End-to-end vLLM serving throughput at the three-epoch budget, averaged over 5 seeds and 9 subsets.
\textbf{Bold} values indicate the top two results.}
\label{tab:vllm-serving}
\end{table}

\section{Per-Subset Greedy Results}
\label{app:subset}

This appendix reports the per-subset acceptance lengths under greedy serving (Table~\ref{tab:subset_greedy}); the probabilistic counterpart is Table~\ref{tab:main} in the main text.

\begin{table*}[t]
\centering
\setlength{\tabcolsep}{2pt}
\begin{tabular}{l c@{\hspace{4pt}} ccccccccc}
\toprule
Method & Avg & HumanEval & GSM8K & NQ & RAG & CNN/DM & BFCL & WMT & MATH-500 & MBPP \\
\midrule
\slk{} & \textbf{4.544} & 5.000 & 5.555 & \textbf{3.533} & \textbf{4.349} & \textbf{4.123} & \textbf{3.665} & \textbf{4.460} & 5.425 & \textbf{4.783} \\
\wtv{} & 4.530 & 5.021 & 5.617 & 3.518 & 4.300 & 4.092 & 3.612 & 4.440 & 5.411 & 4.757 \\
\vsd{} & 4.497 & 4.988 & 5.587 & 3.462 & 4.283 & 3.988 & 3.636 & 4.373 & 5.421 & 4.736 \\
\dblast{} & 4.513 & 5.000 & 5.600 & 3.471 & 4.266 & 4.039 & 3.618 & 4.428 & \textbf{5.469} & 4.731 \\
\ealp{} & 4.515 & 5.002 & 5.582 & 3.482 & 4.292 & 4.062 & 3.611 & 4.396 & 5.453 & 4.751 \\
\ealb{} & \textbf{4.537} & \textbf{5.039} & \textbf{5.638} & 3.504 & 4.284 & 4.066 & 3.651 & 4.420 & 5.465 & 4.766 \\
\bottomrule
\end{tabular}

\caption{Greedy acceptance length after 3 training epochs, averaged over 3 repetitions with the default seed.
Avg is the equal-weight mean over 9 task subsets.
\textbf{Bold} values indicate the top two results in the Avg column and the best result in each subset column.}
\label{tab:subset_greedy}
\end{table*}

\section{Epoch-wise Acceptance Length}
\label{app:epoch}

Table~\ref{tab:epoch-full} reports the complete epoch-wise results under both greedy and probabilistic serving, complementing the probabilistic-only results in Figure~\ref{fig:epoch} in the main text.

\begin{table*}[t]
\centering
\begin{tabular}{l cc cc cc cc}
\toprule
 & \multicolumn{2}{c}{\textbf{epoch 1}} & \multicolumn{2}{c}{\textbf{epoch 2}} & \multicolumn{2}{c}{\textbf{epoch 3}} & \multicolumn{2}{c}{\textbf{epoch 4}} \\
\cmidrule(lr){2-3}\cmidrule(lr){4-5}\cmidrule(lr){6-7}\cmidrule(lr){8-9}
\textbf{method} & greedy & prob. & greedy & prob. & greedy & prob. & greedy & prob. \\
\midrule
\slk{} & \textbf{4.319} & 4.458 & \textbf{4.473} & 4.631 & \textbf{4.544} & 4.740 & \textbf{4.570} & 4.768 \\
\wtv{} & 4.302 & 4.438 & 4.428 & 4.603 & 4.530 & 4.715 & 4.548 & 4.737 \\
\vsd{} & 4.239 & 4.313 & 4.378 & 4.511 & 4.497 & 4.657 & 4.511 & 4.683 \\
\dblast{} & 4.270 & 4.421 & 4.403 & 4.610 & 4.513 & 4.749 & 4.545 & 4.777 \\
\ealp{} & 4.260 & 4.456 & 4.413 & 4.658 & 4.515 & 4.791 & 4.533 & 4.814 \\
\ealb{} & 4.289 & \textbf{4.479} & 4.425 & \textbf{4.660} & 4.537 & \textbf{4.792} & 4.556 & \textbf{4.816} \\
\bottomrule
\end{tabular}

\caption{Acceptance length under greedy and probabilistic serving at the end of each epoch, including a continuation to epoch 4.
Means over 9 subsets, using 5 seeds for probabilistic serving and the default seed for greedy serving.}
\label{tab:epoch-full}
\end{table*}

\section{Comparison with \dpace{}}
\label{app:dpace}

\dpace{} applies dynamic position weighting to the cross-entropy loss rather than optimizing an acceptance-length objective \cite{wu2026dpace}, which is orthogonal to and can be combined with \real{}, and is therefore compared separately from the \EAL{}-class objectives of the main text.
We use $\alpha = 0.5$ for its position weighting.
Table~\ref{tab:dpace-main} compares \dpace{} with \ealp{} and \ealb{}.
Under both serving protocols, \dpace{} is lower than \ealb{} (4.760 vs. 4.792 probabilistic and 4.535 vs. 4.537 greedy).

\begin{table}[t]
\centering
\begin{tabular}{l cc}
\toprule
 & \textbf{greedy} & \textbf{probabilistic} \\
\midrule
\dpace{} & 4.535 & 4.760 \\
\ealp{} & 4.515 & 4.791 \\
\ealb{} & \textbf{4.537} & \textbf{4.792} \\
\bottomrule
\end{tabular}

\caption{Acceptance length after the full 3-epoch budget with \dpace{} included.
Means are equal-weight averages over 9 subsets, using 5 seeds for probabilistic serving and the default seed for greedy serving.
\textbf{Bold} values mark the best result for each protocol.}
\label{tab:dpace-main}
\end{table}

\paragraph{Combination with \dpace{}.}
\dpace{} performs well in Table~\ref{tab:dpace-main} without using the \EAL{} objective, raising the question of whether its position weighting also improves the auxiliary CE term of our \real{} loss.
Table~\ref{tab:dpace} compares the five variants: switching from standard CE to \dpace{}-weighted CE improves probabilistic AL for both \ealp{} and \ealb{}, from 4.791 to 4.797 and from 4.792 to 4.796, respectively, and \ealp{}-\dpace{} attains the best probabilistic AL in the table.
The advantage grows further with a fourth training epoch, with \ealp{}-\dpace{} attaining the highest probabilistic AL (4.838), followed by \ealb{}-\dpace{} (4.823), while \dpace{} reaches 4.784.
This shows that applying \dpace{} weighting to the auxiliary CE term can further improve serving acceptance length.

\begin{table}[t]
\centering
\setlength{\tabcolsep}{1pt}
\begin{tabular}{l cc cc}
\toprule
 & \multicolumn{2}{c}{\textbf{greedy}} & \multicolumn{2}{c}{\textbf{probabilistic}} \\
\cmidrule(lr){2-3}\cmidrule(lr){4-5}
method & epoch 3 & epoch 4 & epoch 3 & epoch 4 \\
\midrule
\dpace{} & 4.535 & 4.555 & 4.760 & 4.784 \\
\ealp{}-CE & 4.515 & 4.533 & 4.791 & 4.814 \\
\ealp{}-\dpace{} & 4.516 & 4.528 & \textbf{4.797} & \textbf{4.838} \\
\ealb{}-CE & 4.537 & \textbf{4.556} & 4.792 & 4.816 \\
\ealb{}-\dpace{} & \textbf{4.538} & 4.554 & 4.796 & 4.823 \\
\bottomrule
\end{tabular}

\caption{Combination of \ealb{} and \ealp{} with \dpace{} position weighting at the three- and four-epoch budgets.
The suffixes -CE and -\dpace{} denote standard CE and \dpace{}-weighted CE, respectively.
\textbf{Bold} values mark the best result in each column.}
\label{tab:dpace}
\end{table}

\section{Pilot Experiments}
\label{app:toy}

This appendix reports the small-scale predecessor of the main study (Qwen3-8B target, Qwen3-0.6B draft, GSM8K prompts for training and 64 held-out prompts for evaluation), whose results in Table~\ref{tab:toy} shaped the design of the main-study objectives and the auxiliary loss.

\subsection{Setup}

We use Qwen3-8B as the frozen target and fully fine-tune Qwen3-0.6B as the draft \cite{yang2025qwen3} on GSM8K prompts, with window $\gamma = 7$ and temperature 1 using the full softmax.
Training windows contain 1024 tokens, and evaluation uses 64 held-out prompts.
We use pure objectives based on $\widehat{J}_{\textsc{Branch}}$ and $\widehat{J}_{\textsc{Path}}$ without auxiliary CE, corresponding to \ealb{} w/o CE and \ealb{} w/o CE\&RB, respectively.
We compare seven objectives: CE, single-step TV, LK$^\lambda$ ($\eta = 3$), \wtv{}, $\widehat{J}_{\textsc{Branch}}$, $\widehat{J}_{\textsc{Path}}$, and \wtv{} trained on greedy decodes (the original \wtv{} data recipe).
The pilot \wtv{} objectives also omit auxiliary CE.
Two ablations add an auxiliary KL term to the losses based on $\widehat{J}_{\textsc{Branch}}$ and $\widehat{J}_{\textsc{Path}}$, and a third detaches the prefix weight $W$ in $\widehat{J}_{\textsc{Branch}}$.
For variants with an auxiliary KL term, the total loss is $1-\widehat J/\gamma+L_{\mathrm{KL}}$, with the auxiliary coefficient set to 1.
\subsection{Results}

\begin{table}[t]
\centering
\begin{tabular}{l c}
\toprule
\textbf{Objective} & \textbf{Acceptance length (T=1)} \\
\midrule
Base draft & 2.5833 \\
CE (KL) & 3.1354 \\
TV ($\gamma{=}1$) & 2.3750 \\
LK$^\lambda$ & 3.6406 \\
\wtv{} & 3.9375 \\
$\widehat{J}_{\textsc{Branch}}$ & 3.5312 \\
$\widehat{J}_{\textsc{Path}}$ & 0.5052 \\
\wtv{} on greedy data & 2.5000 \\
\midrule
\multicolumn{2}{l}{\textit{Ablations}} \\
$\widehat{J}_{\textsc{Branch}}$ with KL & 3.6302 \\
$\widehat{J}_{\textsc{Path}}$ with KL & 3.1458 \\
$\widehat{J}_{\textsc{Branch}}$, detach $W$ & 1.5573 \\
\bottomrule
\end{tabular}

\caption{Pilot study: verifier acceptance length at T=1 on 64 held-out GSM8K prompts.}
\label{tab:toy}
\end{table}

The following findings shaped the main study.
\begin{enumerate}[leftmargin=1.4em,itemsep=1pt]
\item \textbf{The exact objective does not significantly outperform the surrogate in this setting.}
In this small-scale experiment, per-pair correlations are small, evaluation variance is large, and training is limited, so the comparison between training with $\widehat{J}_{\textsc{Branch}}$ and \wtv{} remains inconclusive.
This provides a complementary setting in which exact estimation does not yield a clear serving advantage.
\item \textbf{The path objective without CE is fragile.}
It collapsed in all four training runs through a positive feedback loop in which $b \to 0$ concentrates the loss on surviving positions, and the path estimator has higher gradient variance than both \wtv{} and the branch estimator.
Training with $\widehat{J}_{\textsc{Branch}}$ outperforms training with $\widehat{J}_{\textsc{Path}}$ by 3.03 AL on average, a large and statistically significant advantage.
\item \textbf{Detaching $W$ fails.}
The detached variant scores 1.56, supporting the role of the prefix-weight gradient in Equation~\eqref{eq:grad} in stabilizing training.
\item \textbf{The auxiliary loss stabilizes training.}
Across four independent training runs, training with $\widehat{J}_{\textsc{Branch}}$ without KL succeeds only once, reaching 3.53 serving AL in Table~\ref{tab:toy}.
Using $\widehat{J}_{\textsc{Branch}}$ with KL stabilizes training and reaches 3.63 serving AL, but does not significantly outperform training with either $\widehat{J}_{\textsc{Branch}}$ without KL or \wtv{}.
This motivates the auxiliary CE term of the main study.
\item \textbf{The data recipe matters.}
\wtv{} scores 2.50 when trained on greedy decodes and 3.94 when trained on sampled continuations, supporting the requirement that real acceptance objectives train on target-sampled trajectories.
\end{enumerate}

\section{Serving-Variance Analysis and Regularization Attempts}
\label{app:variance}

Although \real{} improves mean acceptance length, we still observe substantial round-to-round variability during inference, with acceptance lengths rising and falling across consecutive rounds.
We explore two extensions of \ealb{} to reduce this variability and improve acceptance at early positions.

\paragraph{Variance penalty.}
Within-request fluctuations account for about 89\% of total serving variance in our evaluation (Table~\ref{tab:stage2}), motivating \ealv{}, which maximizes a mean--variance objective:
\begin{equation}
J_{\mathrm V}=\mu-\lambda_{\mathrm{var}}(M_2-\mu^2).
\end{equation}
Here $M_2=\sum_k(2k-1)S_k$ is the second moment of the accepted draft-token count, $\mu$ is its mean, and $S_k=P(L\ge k)$, so $M_2-\mu^2$ measures variance across decoding rounds.

\paragraph{Prefix objective.}
Separately, \ealb{} and \ealp{} have weaker retention at the first position than static LK, while very small acceptance probabilities at later positions may provide insufficient training signal.
We therefore test \ealk{}, which maximizes acceptance length capped at a fixed budget $K\leq\gamma$:
\begin{equation}
J_K=\mathbb{E}[\min(L,K)]=\sum_{k=1}^{K}S_k.
\end{equation}
Setting $K=\gamma$ recovers the full-window objective $J_\gamma$ in Equation~\eqref{eq:tailsum}.

\paragraph{Design.}
For each variant, the corresponding training estimate replaces $\widehat{J}_{\textsc{Branch}}$ in Equation~\eqref{eq:real-loss}, retaining the fixed denominator $\gamma$ and the auxiliary CE term.
Starting from the \ealb{} checkpoint after two epochs, we continue training with these two variants using hot continuation, resuming the optimizer, scheduler, and training state from the checkpoint and setting $K=5$ and $\lambda_{\mathrm{var}}=0.05$.
The training-time mean $\mu$ is the average of the per-block branch \EAL{} estimates $\widehat{J}_{\textsc{Branch}}=\sum_{k=1}^{\gamma}W_{k-1}a_k$ over valid blocks, where the variance penalty is computed within each training sequence and averaged equally across sequences.
We also include a mean-preservation guard $(\lambda\rho/\gamma)[\mu_{\min}-\mu]_+$ in the total loss, where $\mu_{\min}=\mu_0-\delta$, $\mu_0$ is the mean of $\widehat{J}_{\textsc{Branch}}$ at the end of the \ealb{} run, $\delta=0.05$, and $\rho=0.9$, activated only when $\mu<\mu_{\min}$.
This regularization strength is approximately $6.5\lambda_{\mathrm{var,ref}}$, where $\lambda_{\mathrm{var,ref}}=\delta/V_0\approx0.0077$ and $V_0$ is the measured within-request variance.

\begin{table}[t]
\centering
\setlength{\tabcolsep}{1pt}
\begin{tabular}{@{}l cccc@{}}
\toprule
\textbf{Metric} & \wtv{} & \ealb{} & \ealv{} & \ealk{} \\
\midrule
Validation $\widehat{J}_{\textsc{Branch}}$ & 4.3570 & 4.3795 & 4.3790 & 4.3669 \\
Within var. (greedy) & 6.499 & 6.554 & 6.548 & 6.526 \\
Within var. (prob.) & 6.252 & 6.340 & 6.343 & 6.312 \\
Between var. (greedy) & 0.804 & 0.825 & 0.830 & 0.818 \\
Between var. (prob.) & 0.878 & 0.890 & 0.890 & 0.839 \\
Max streak (greedy) & 50 & 26 & 31 & 23 \\
Max streak (prob.) & 33 & 20 & 31 & 17 \\
\bottomrule
\end{tabular}

\caption{Outcomes after one additional training epoch with hot continuation, with \wtv{} and \ealb{} as standard three-epoch references. All methods include a CE term.}
\label{tab:stage2}
\end{table}

\begin{figure}[t]
\centering
\subcaptionbox{Greedy serving\label{fig:stage2-hist-greedy}}{\includegraphics[width=0.49\columnwidth]{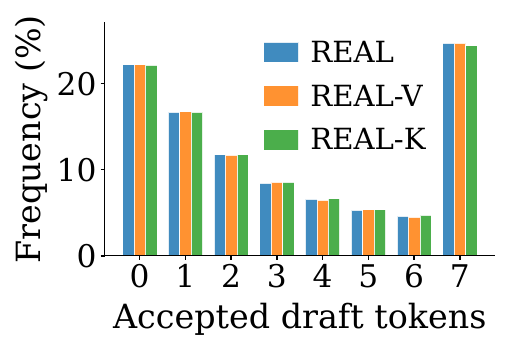}}\hfill
\subcaptionbox{Probabilistic serving\label{fig:stage2-hist-prob}}{\includegraphics[width=0.49\columnwidth]{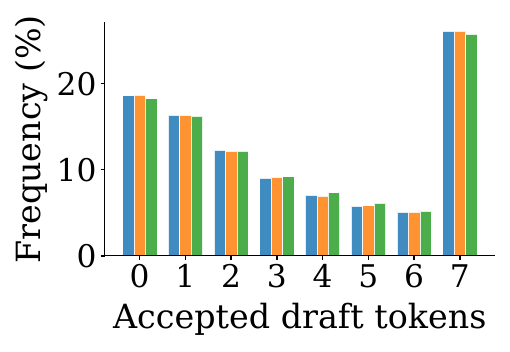}}
\caption{Distributions of accepted draft-token counts per decoding round for \ealb{}, \ealv{}, and \ealk{}.
All three methods show similar distributions under both serving protocols.}
\label{fig:stage2-hist}
\end{figure}

\begin{table}[t]
\centering
\begin{tabular}{c cc}
\toprule
$K$ & \ealb{} & \ealk{} \\
\midrule
1 & 0.81 & 0.82 \\
2 & 1.46 & 1.47 \\
3 & 1.99 & 2.01 \\
4 & 2.43 & 2.45 \\
5 & 2.80 & 2.82 \\
6 & 3.11 & 3.13 \\
7 & 3.37 & 3.39 \\
\bottomrule
\end{tabular}

\caption{Probabilistic $J_K$ after hot continuation, evaluated at each cutoff $K=1,\dots,7$.
\ealk{} is trained with $K=5$. All methods include a CE term.}
\label{tab:ealk-jk}
\end{table}

\paragraph{Outcome.}
The validation means of $\widehat{J}_{\textsc{Branch}}$ for \ealv{} and \ealk{} remain close to that for \ealb{} (Table~\ref{tab:stage2}).

\ealv{} shows no consistent reduction in within-request variance under either greedy or probabilistic serving at $\lambda_{\mathrm{var}}=0.05$ (Table~\ref{tab:stage2}).
The per-round histograms are nearly identical across the three objectives, decreasing monotonically from 0 to 6 accepted draft tokens but rising sharply at 7, indicating a substantial full-acceptance tail (Figure~\ref{fig:stage2-hist}).

\ealk{} reduces the max streak, defined as the longest run of consecutive rounds with at most one accepted draft token and maximized over requests, to 23 rounds under greedy serving and 17 rounds under probabilistic serving, compared with 26 and 20 for \ealb{} (Table~\ref{tab:stage2}).
\wtv{} also exhibits much longer low-acceptance streaks than \ealb{}, reaching 50 versus 26 rounds under greedy serving and 33 versus 20 under probabilistic serving (Table~\ref{tab:stage2}).
As an auxiliary diagnostic, \ealk{} increases probabilistic $J_K$ by 0.01 to 0.02 over \ealb{}, indicating the intended early-position shift (Table~\ref{tab:ealk-jk}).
However, this small auxiliary gain does not carry over to the official serving evaluation, where \ealk{} has lower AL than \ealb{}.

We leave further exploration of variance-aware objectives and regularization strategies to future work.

\end{document}